\documentclass[sigconf,natbib=true]{acmart}

\AtBeginDocument{%
  }

\usepackage{amsmath,graphicx}
\usepackage{paralist}
\usepackage{booktabs}
 \usepackage{ulem}
\usepackage{accents}
\usepackage{multirow}
\definecolor{brightmaroon}{rgb}{0.76, 0.13, 0.28}
\definecolor{softmagenta}{RGB}{219,112,147}

\usepackage[english]{babel}

\usepackage{enumitem}
\usepackage{paralist}
\usepackage{arydshln}
\usepackage{soul}
\usepackage{ulem}
\definecolor{cyan}{RGB}{0,255,255}
\definecolor{lightcyan}{RGB}{21,255,255}
\definecolor{palecyan}{RGB}{220,255,255}
\definecolor{magenta}{RGB}{255,0,255}
\definecolor{palepink}{RGB}{203, 153, 197}
\definecolor{lightblue}{RGB}{109, 191, 235}

\copyrightyear{2025}
\acmYear{2025}
\setcopyright{acmlicensed}\acmConference[SIGIR '25]{Proceedings of the 48th International ACM SIGIR Conference on Research and Development in Information Retrieval}{July 13--18, 2025}{Padua, Italy}
\acmBooktitle{Proceedings of the 48th International ACM SIGIR Conference on Research and Development in Information Retrieval (SIGIR '25), July 13--18, 2025, Padua, Italy}
\acmDOI{10.1145/3726302.3729884}
\acmISBN{979-8-4007-1592-1/2025/07}

\begin{document}

\title{A Unified Retrieval Framework with Document Ranking and EDU Filtering for Multi-document Summarization}

\author{Shiyin Tan}
\orcid{0000-0001-8316-2838}
\authornote{Both authors contributed equally to this research.}
\affiliation{%
  \institution{Institute of Science Tokyo}
  \city{Tokyo}
  \country{Japan}
}
\email{tanshiyin1107@gmail.com}

\author{Jaeeon Park}
\authornotemark[1]
\orcid{0000-0002-0410-6457}
\affiliation{%
  \institution{Institute of Science Tokyo}
  \city{Tokyo}
  \country{Japan}
}
\email{jaeeon@lr.pi.titech.ac.jp}

\author{Dongyuan Li}
\authornote{Corresponding author.}
\orcid{0000-0002-4462-3563}
\affiliation{%
  \institution{The University of Tokyo}
  \department{Center for Spatial Information Science}
  \city{Tokyo}
  \country{Japan}
}
\affiliation{%
  \institution{Institute of Science Tokyo}
  \city{Tokyo}
  \country{Japan}
}
\email{lidy94805@gmail.com}

\author{Renhe Jiang}
\orcid{0000-0003-2593-4638}
\affiliation{%
  \institution{The University of Tokyo}
  \department{Center for Spatial Information Science}
  \city{Tokyo}
  \country{Japan}
}
\email{jiangrh@csis.u-tokyo.ac.jp}

\author{Manabu Okumura}
\orcid{0009-0001-7730-1536}
\affiliation{%
  \institution{Institute of Science Tokyo}
  \city{Tokyo}
  \country{Japan}
}
\email{oku@pi.titech.ac.jp}
\renewcommand{\shortauthors}{Shiyin Tan, Jaeeon Park, Dongyuan Li, Renhe Jiang, \& Manabu Okumura.}

\begin{abstract}

In the field of multi-document summarization (MDS), transformer-based models have demonstrated remarkable success, yet they suffer 
an input length limitation. 
Current methods apply truncation after the retrieval process to fit the context length; however, they heavily depend on manually well-crafted queries, which are impractical to create for each document set for MDS. 
Additionally, these methods retrieve information at a coarse granularity, leading to the inclusion of irrelevant content.
To address these issues, we propose a novel retrieval-based framework that integrates query selection and document ranking and shortening 
into a unified process. 
Our approach identifies the most salient elementary discourse units (EDUs) from input documents and utilizes them as latent queries. These queries guide the document ranking by calculating relevance scores.
Instead of 
traditional truncation, our approach filters out irrelevant EDUs to fit the context length, ensuring that only critical information is preserved for summarization.
We evaluate our framework on multiple MDS datasets, demonstrating consistent improvements in ROUGE metrics while confirming its scalability and flexibility across diverse model architectures. Additionally, we validate its effectiveness through an in-depth analysis, emphasizing its ability to dynamically select appropriate queries and accurately rank documents based on their relevance scores. These results demonstrate that our framework effectively addresses context-length constraints, establishing it as a robust and reliable solution for MDS.\footnote{Our code is available at \textcolor{blue}{\url{https://github.com/ShiyinTan/ReREF}}.}


\end{abstract}




\maketitle

\section{Introduction}
\label{sec:intro}

Multi-document summarization (MDS) is a task that aims to generate concise and coherent summaries by synthesizing information from multiple documents on the same topic \cite{li-etal-2020-leveraging-graph, jin-etal-2020-multi, mao-etal-2020-multi, DBLP:conf/acl/PangLTY21, DBLP:conf/sigir/Ma21}. MDS can lead to diverse applications, such as  
news aggregation~\cite{fabbri-etal-2019-multi,DBLP:conf/sigir/KhatuyaSG0024,DBLP:conf/sigir/ChenLGCZ00024}, 
scientific research \cite{lu-etal-2020-multi-xscience, deyoung-etal-2021-ms,DBLP:conf/sigir/Wang0LLTW24}, 
and legal document analysis \cite{DBLP:conf/sigir/Malik0FRC24,1cd5ffa600784f008953105ce249109d,DBLP:conf/nips/GuhaNHRCKCPWRZT23}. 
Current MDS approaches can be categorized into two classes: Graph-based models~\cite{DBLP:conf/naacl/PasunuruLBRD21,DBLP:conf/eacl/ZhangEDSJB23,DBLP:conf/aaai/Li0L23, DBLP:conf/emnlp/0006H21, DBLP:conf/www/Qu24} and pre-trained language models~\cite{beltagy2020longformer,DBLP:conf/acl/XiaoBCC22,DBLP:conf/acl/PuduppullyJCS23}. 
Graph-based models rely on auxiliary information (\textit{e.g.,} discourse structures) as an input graph to capture the cross-document relationships, while pre-trained language models use the attention mechanisms to capture them. 

\begin{figure}[h]
	\centering
        \includegraphics[scale=0.5]{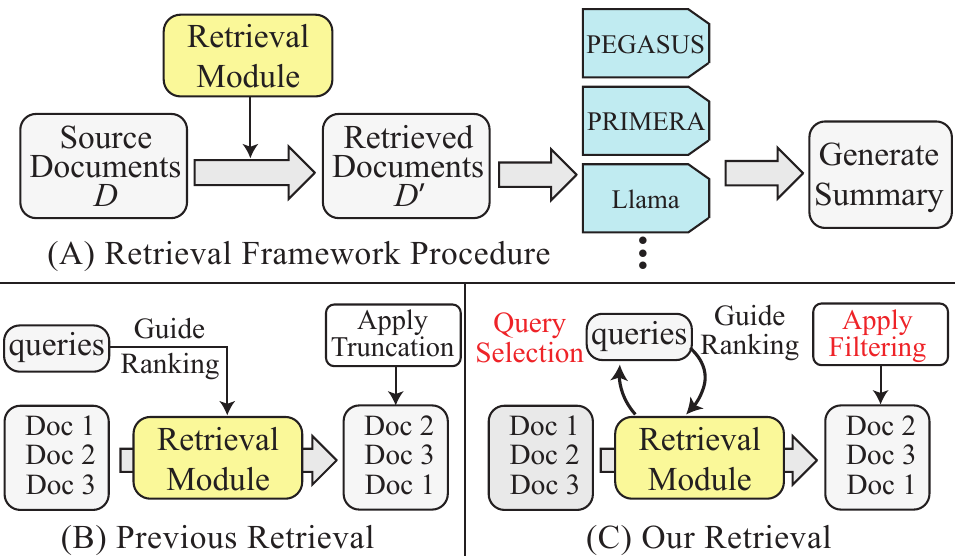}
    \vspace{-5pt}
	\caption{(A) The overall procedure of our retrieval model, which is model agnostic. {(B) Previous retrieval needs queries as inputs and ranking over documents. (C) Our retrieval automatically selects queries for document ranking and applies filtering over EDUs}.}
    \label{fig:procedure}
    \vspace{-10pt}
\end{figure}

All these summarization models employ transformer~\cite{10.5555/3295222.3295349} as a text encoder and are consequently constrained by a fixed input length, limiting the number of tokens they can process.
A common solution is to apply truncation, dropping the last tokens of input documents to fit within the context length.
However, this naive approach risks discarding critical information 
for summarization~\cite{10.1145/3529754,wolhandler-etal-2022-multi}. 
Thus, \textit{how to retain sufficient critical information within the length limitation} has become a crucial issue for MDS to enhance the quality of summaries.
{Current approaches often follow a ``retrieve-then-summarize'' paradigm to alleviate this issue~\cite{DBLP:journals/corr/abs-2109-07943,DBLP:journals/corr/abs-2406-12494,DBLP:conf/emnlp/GiorgiSWBLWC23,zhang2021tackling}, as shown in Figure~\textcolor{blue}{\ref{fig:procedure}} (A). They use manually created \textbf{queries} as guidance to rank \textbf{documents or sentences} from input documents or external knowledge bases and retrieve the top ranked contents to generate summaries. 
For instance, LightPAL~\cite{DBLP:journals/corr/abs-2406-12494} and DYLE~\cite{DBLP:conf/acl/MaoWN0Z0DZAR22} use dataset-provided queries to retrieve passages and sentences, respectively, for summarization}.

Although these frameworks effectively retrieve documents from a large document collection, they face two major issues in MDS, {as shown in Figure~\textcolor{blue}{\ref{fig:procedure}} (B)}: 
\textbf{(i) 
These retrieval methods~\cite{DBLP:conf/emnlp/GiorgiSWBLWC23} often heavily depend on manually well-crafted queries} to guide the ranking of passages or documents. For example, queries require precise human-written topic statements to ensure accurate ranking. However, MDS datasets do not provide the corresponding queries, {and it is impractical to create them for each document set for MDS,  
as creating a topic statement requires reading the entire document set}. 
Additionally, they overlook the fact that documents in MDS often cover the same topic from different perspectives~\cite{lin2002single}, {which suggests that queries can be directly inferred from documents}.
\textbf{(ii) 
Current retrieval models focus on a coarse granularity unit}, which inevitably includes irrelevant information for summarization, as pointed out in \cite{DBLP:conf/emnlp/Chen0C0MZ0024,DBLP:conf/icml/ShiCMSDCSZ23,DBLP:conf/emnlp/00110PCML0024}. 
Specifically, some retrieval methods~\cite{DBLP:journals/corr/abs-2406-12494,DBLP:journals/corr/abs-2403-19889,DBLP:conf/sigir/RajapakseYR24} focus on passages as retrieval units, while others~\cite{DBLP:conf/acl/MaoWN0Z0DZAR22,DBLP:conf/emnlp/KwonKKO21,DBLP:conf/sigir/HarelTSK22} focus on sentences. 

To address the {above issues}, we propose a model-agnostic \underline{Re}trieval framework, called ReREF, which unifies document \underline{R}anking and \underline{E}lementary discourse unit (EDU)\footnote{{EDUs~\cite{carlson2003building} are minimal coherent segments of text and are usually smaller than sentences. More details about EDUs are introduced in Section~\textcolor{blue}{\ref{sec:problem_edu}}}.} \underline{F}iltering for MDS. The overall procedure of ReREF is shown in {Figure~\textcolor{blue}{\ref{fig:procedure}} (C)}. 
Firstly, to automatically select queries, we treat the problem as EDU ranking, where the top-ranked EDUs indicate {that} they have high relevance to the reference summary and can be used as queries for document ranking. 
Secondly, to fully utilize the limited length with critical information and alleviate the distraction from extraneous details, we not only rank documents to ensure that important ones are not truncated but also filter out bottom-ranked EDUs to eliminate irrelevant segments in sentences. 
Then, we employ the Expectation-Maximization (EM) algorithm \cite{DEMP1977} to unify automatic query selection, document ranking, and EDU filtering into a cohesive process. 

We apply ReREF to seven summarizers on four MDS datasets, using both fully supervised and few-shot settings. The results demonstrate that ReREF consistently improves summarization quality {from the perspective of ROUGE-scores} over the original summarizers. Additionally, human evaluations further assess the quality of the generated summaries, showing improvements in both informativeness and fluency with ReREF.
The main contributions of this study can be summarized as follows: 
\begin{itemize}[left=0em]
\item We {design} automatic query selection and irrelevant information filtering at the EDU level to avoid the need of human-written queries and reduce the space consumption of irrelevant {content}.
\item We unify automatic query selection, document ranking, and EDU filtering into a model-agnostic retrieval framework using the 
EM algorithm to address the context length limitation.
\item We evaluate ReREF with various summarizers in both fully supervised and few-shot settings to demonstrate its effectiveness. 
Additional human evaluations show that ReREF improves the quality of the generated summaries.
\end{itemize}

\section{Related Work}

\subsection{Multi-document Summarization}

MDS is a task that summarizes information from a set of related documents on the same topic~\cite{DBLP:conf/sigir/Wang0LLTW24}, which can be categorized into {three} classes: 
{Statistical and linguistic feature-based methods, }
graph-based models, and  pre-trained language models. 
{\textbf{Statistical and linguistic feature-based methods} are generally unsupervised and do not require labeled training data. Typical techniques include scoring sentences~\cite{goldstein2000multi} based on term frequency inverse document frequency (TF-IDF), keyword occurrence, and semantic similarity measures~\cite{ozatecs2016sentence}.}
\textbf{Graph-based models} require auxiliary information (\textit{e.g.,} discourse structures) to build an input graph to capture the cross-document relationships. For example, models like BartGraphSum \cite{DBLP:conf/naacl/PasunuruLBRD21} and IESum~\cite{DBLP:conf/eacl/ZhangEDSJB23} utilize graph representations, constructed by Information Extraction (IE), to encode complex relationships within and across documents. HGSUM \cite{DBLP:conf/aaai/Li0L23} extends this concept by employing heterogeneous graphs to represent words, sentences, and documents at multiple semantic levels, further improving abstractive summarization performance.
\textbf{Pre-trained language models} follow a transformer-based encoder-decoder architecture to capture the cross-document relationships by attention mechanisms.
Efficient pre-trained transformers that can process long sequences (\textit{e.g.,} Longformer~\cite{beltagy2020longformer}) have been also proven successful in summarization, typically by the ability to process long inputs, connecting information across the entire sequence.
PRIMERA \cite{DBLP:conf/acl/XiaoBCC22} and Centrum~\cite{DBLP:conf/acl/PuduppullyJCS23} introduce a specialized architecture for MDS, utilizing Longformer’s global attention mechanisms to capture key information from multiple documents.
Both the graph-based models and the pre-trained language models are well designed to capture key information across documents, but they still face challenges brought by context-length constraints. 
Specifically, they simply apply a truncation strategy that drops the last tokens in documents to fit the context length, leading to the lose of critical information necessary for summarization.
ReREF builds upon such a challenge by introducing a retrieval {framework} that integrates both ranking and filtering into a unified process. 

\subsection{Retrieval Frameworks for Summarization}
The studies on MDS 
often adopt a ``retrieve-then-summarize'' paradigm~\cite{DBLP:journals/corr/abs-2109-07943,DBLP:journals/corr/abs-2406-12494,DBLP:conf/emnlp/GiorgiSWBLWC23,zhang2021tackling}, consisting of two main modules: a \textit{Retriever} and a \textit{Summarizer}. 
The \textit{Retriever} selects passages based on the relevance between queries (either provided by humans or available in datasets) and passages (sourced from input documents or external knowledge bases). 
The \textit{Summarizer} then generates summaries using the retrieved passages.
For instance, LightPAL~\cite{DBLP:journals/corr/abs-2406-12494} implements a two-step retrieval process: initially retrieving passages with a conventional retriever (\textit{e.g.,} BM25~\cite{10.1561/1500000019}), followed by further refinement using large language models (LLMs). LogicSumm~\cite{DBLP:journals/corr/abs-2403-19889} utilizes LLMs for both document retrieval and summarization. 
DYLE~\cite{DBLP:conf/acl/MaoWN0Z0DZAR22} employs RoBERTa~\cite{DBLP:journals/corr/abs-1907-11692} to retrieve sentences from input documents.
Despite effectively retrieving passages from a large document collection, these frameworks heavily depend on well-crafted queries~\cite{DBLP:conf/emnlp/GiorgiSWBLWC23} (\textit{e.g.,} topic statements) to guide ranking and often use sentences or passages as their smallest retrieval units.
However, in 
MDS, where documents typically cover the same topic from different perspectives~\cite{lin2002single}, queries can often be inferred directly from the documents as they share critical information. Moreover, sentences and paragraphs frequently contain a mix of relevant and irrelevant information, and retrieval at this level inevitably includes irrelevant information for summarization. 
Our framework addresses these challenges by automatically generating queries from documents to avoid manual query creation and by filtering at the level of EDUs to ensure critical information is captured within the summarizer's limited context length.


\begin{figure}[h]
	\centering
        \includegraphics[scale=0.38]{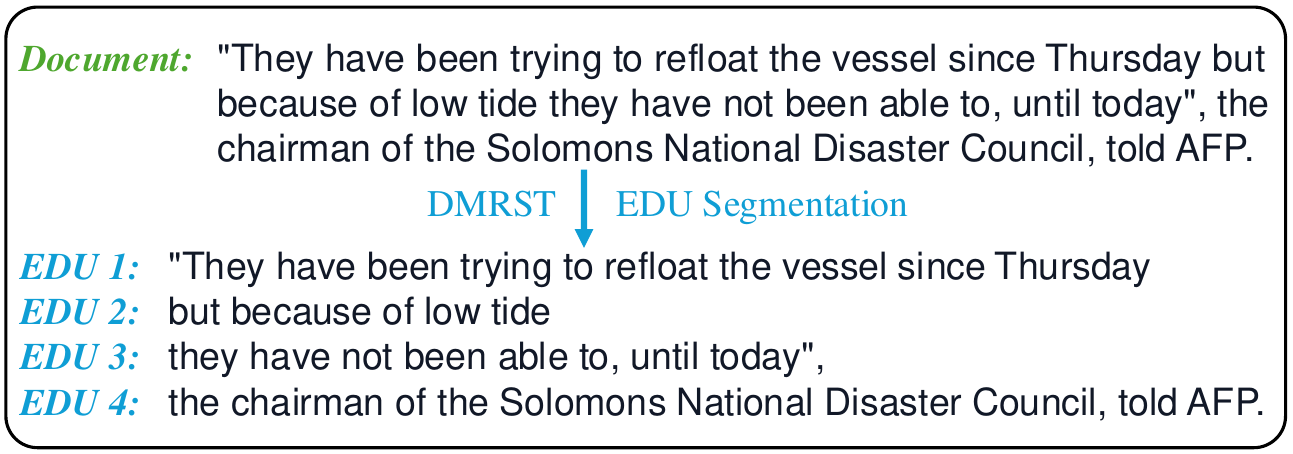}
    \vspace{-5pt}
	\caption{Example EDUs, {segmented by DMRST~\cite{liu-etal-2021-dmrst}}, demonstrate their finer granularity 
    than sentences while being more semantically meaningful than words or tokens.}
    \vspace{-10pt}
    \label{fig:example_segmentation}
\end{figure}

\section{Preliminaries}

\subsection{Notations}
In this paper, we use capital and calligraphic letters to represent a document and a set of documents, respectively. For instance, $D_i$ denotes the text of document $i$, while $\mathcal{D}=\{D_1, \dots, D_n\}$ represents a collection of documents. 
Bold lowercase and uppercase letters are used to denote vectors and matrices, respectively. For example, $\mathbf{d}_i$ represents the embedding of document $i$, and $\mathbf{D}$ refers to all document embeddings in a document set $\mathcal{D}$. 
Additionally, we let $T$, $E$ and $D$ denote tokens, EDUs, and documents, respectively.

\subsection{Problem Statement}


MDS is defined as a generation task, given a set of documents on the same topic, 
where the objective is to generate a summary $S^*$ that accurately and coherently represents the information in $\mathcal{D}$. 
Models for this task are typically trained in a supervised manner to minimize the difference between $S^*$ and a reference summary.

However, MDS is inherently challenging due to the long-context nature of its inputs, often exceeding the input length limitation of summarization models. 
To address this limitation, many approaches~\cite{DBLP:journals/corr/abs-2109-07943,DBLP:conf/emnlp/ShiGZCCRR23,DBLP:conf/emnlp/GiorgiSWBLWC23,DBLP:journals/corr/abs-2406-12494, DBLP:conf/acl/LiWLC18} adopt a ``retrieve-then-summarize'' paradigm. 
{In this study, we follow 
this
paradigm and focus on designing a retrieval framework to enhance the quality of summary for MDS}. 



\subsection{EDU Segmentation}
\label{sec:problem_edu}
EDUs~\cite{carlson2003building}, derived from Rhetorical Structure Theory (RST)~\cite{mann1988rhetorical}, represent the minimal, coherent segments of discourse, that encapsulate individual propositions or ideas. 
A key aspect of our framework is leveraging EDUs to capture the semantic and structural granularity of input documents while filtering out irrelevant EDUs. This ensures that critical information is retained while irrelevant content is effectively filtered out.
However, standard MDS datasets often lack EDU segmentation annotations. 
To address this limitation, we utilize the DMRST parser~\cite{liu-etal-2021-dmrst}, a reliable tool designed for document-level segmentation. The DMRST parser
offering robust multilingual support and delivering high-quality EDU segmentation.
An example of EDU segmentation is shown in Figure~\textcolor{blue}{\ref{fig:example_segmentation}}.

\begin{figure*}[h]
	\centering
        \includegraphics[scale=0.72]{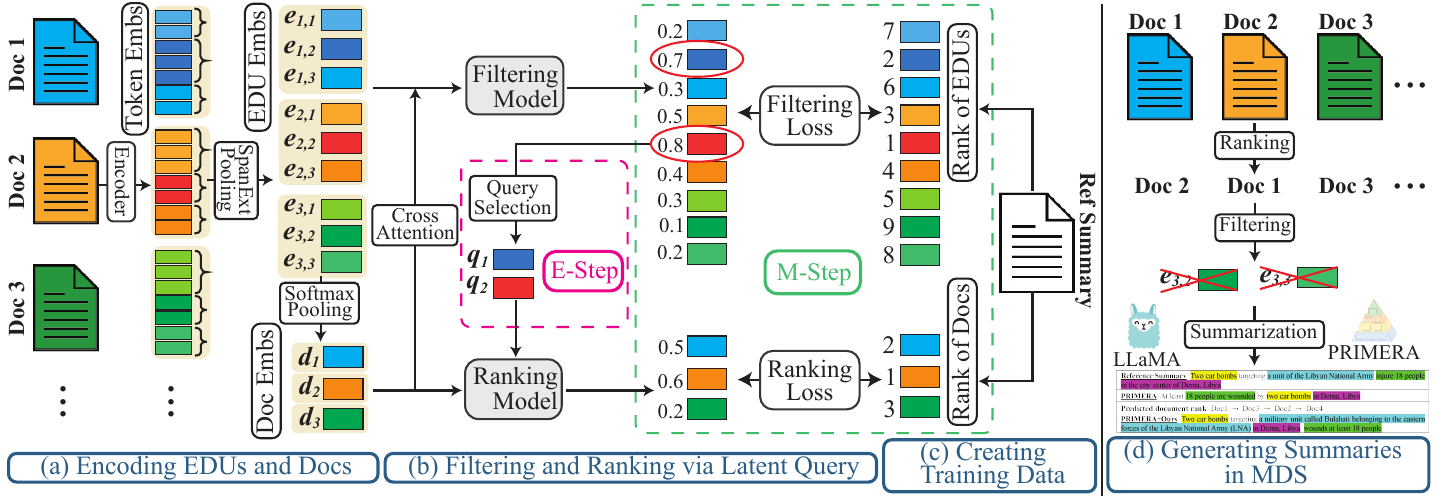}
    \vspace{-5pt}
	\caption{{The overall process} of {ReREF}, which extracts latent queries and performs ranking and filtering  simultaneously.}
    \label{fig:framework}
    \vspace{-5pt}
\end{figure*}

\section{{Methodology}}
In this section, we introduce our unified retrieval framework of ranking and filtering for MDS.  
First, we describe how EDU and document embeddings are encoded and aggregated through pooling techniques (Section~\textcolor{blue}{\ref{sec:method_2}}). 
Next, to automatically select latent queries and simultaneously perform ranking and filtering, we leverage the EM algorithm, which dynamically filters irrelevant content and ranks documents based on their relevance (Section~\textcolor{blue}{\ref{sec:method_3}}).
Then, we present the training dataset creation for our retrieval framework, which includes EDU-scoring and document-scoring (Section~\textcolor{blue}{\ref{sec:method_1}}).
Finally, we introduce the process of integrating our retrieval framework with the truncation strategy during inference and summarizing for the retrieved datasets with a summarizer (Section~\textcolor{blue}{\ref{sec:method_4}}). 



\subsection{Encoding EDUs and Documents}
\label{sec:method_2}
{As shown in Figure~\textcolor{blue}{\ref{fig:framework}} (a)}, we first encode EDUs and documents into suitable representations.
We encode input documents to token embeddings and apply a pooling strategy over the token embeddings to get the EDU and document representations. 
As we introduced in Section~\textcolor{blue}{\ref{sec:problem_edu}}, the EDUs are obtained by the EDU segmentation.

\paragraph{\textbf{Token Embeddings}}
Given a set of input documents $\mathcal{D}$, we encode each document $D_i$ into token embeddings $\mathbf{T}_i$ by  Longformer~\cite{beltagy2020longformer}. Since the document $D_i$ may still meet the context limitation of Longformer, we apply a chunking strategy by splitting $D_i$ into chunks with a fixed size, ensuring that all tokens are processed within the model’s input constraints: 
\begin{align}
    \mathbf{T}_{i}^j &= \text{Enc}(C_i^{j}), \\
    \mathbf{T}_i &= \text{Concat}(\mathbf{T}_{i}^0, ..., \mathbf{T}_{i}^{|C_i|}),
\end{align}
where $C_i^{j}$ represents the $j$-th chunk for document $D_i$, $|C_i|$ denotes the total number of chunks in $D_i$, and $\mathbf{T}_i^{j}$ corresponds to the token embeddings for the $j$-th chunk. All tokens in document $D_i$ are concatenated into token embeddings $\mathbf{T}_i$, which can be pooled into EDU and document embeddings. 

\paragraph{\textbf{EDU Embeddings}} 
After obtaining all token embeddings, EDUs can be derived using a pooling function, such as mean pooling. In this work, we adopt an attention-based weighted pooling method, SpanExt~\cite{lee-etal-2017-end}, which leverages a self-attentive span representation to effectively capture both the internal structure and contextual information of text spans.
For EDU $E_{i,j}$, which denotes the $j$-th EDU in document $D_i$, its embedding is computed as:
\begin{equation}
     \mathbf{e}_{i,j} = \sum_{k=1}^{\ell_{i,j}} {a}_{(i,j,k)} \cdot \mathbf{T}_{(i,j,k)},
     \label{eq:edu_encoding}
\end{equation}
where $\ell_{i,j}$ indicates the total token number of EDU $E_{i,j}$ and $\mathbf{T}_{(i,j,k)}$ denotes the $k$-th token embedding of EDU $E_{i,j}$. The pooling weights ${a}_{(i,j,k)}$ are computed as:
\begin{align}
    {a}_{(i,j,k)} &= \frac{\exp(\alpha_{(i,j,k)})}{\sum_{l=1}^{\ell_{i,j}} \exp(\alpha_{(i,j,l)})}, \\[0.4em]
     \alpha_{(i,j,k)} &= \mathbf{W}_2 \cdot \text{ReLU}(\mathbf{W}_1 \cdot \mathbf{T}_{(i,j,k)} + \mathbf{b}_1) + \mathbf{b}_2,
\end{align}
where $\alpha_{(i,j,k)}$ is the unnormalized attention score, and $\mathbf{W}_1$, $\mathbf{W}_2$, $\mathbf{b}_1$, $\mathbf{b}_2$ are trainable parameters. 
The normalized attention ${a}_{(i,j,k)}$ 
reflects the relative contribution of each token within all the tokens in the span of EDU $E_{i,j}$. The resulting EDU embeddings capture text semantics at a finer granularity, providing a robust foundation for subsequent filtering and ranking processes.


\paragraph{\textbf{Document Embeddings}}
To obtain a holistic representation for each document, we aggregate the embeddings of its EDUs into a single document embedding using a gated pooling mechanism \cite{716791}. The process begins by transforming the EDU embeddings through a non-linear mapping function $g(\cdot)$, implemented by Feedforward Neural Network (FFN) in our approach.
We then aggregate the transformed EDU embeddings
using a Softmax Gating function~\cite{DBLP:conf/iclr/ShazeerMMDLHD17}:
\begin{align}
    \mathbf{d}_i = \sum_{j=1}^{m_i} \beta_{i,j} \cdot g(\mathbf{e}_{i,j}), \ \ \  \beta_{i,j} = \frac{\exp(\mathbf{w}^\top g(\mathbf{e}_{i,j}))}{\sum_{k=1}^{m_i} \exp(\mathbf{w}^\top g(\mathbf{e}_{i,k}))},
    \label{eq:document_embedding_pooling}
\end{align}
where $m_i$ is the total EDU number of document $D_i$, $\mathbf{w} \in \mathbb{R}^d$ is a trainable vector serving as a gating parameter, and $\beta_{i,j}$ represents the pooling weight assigned to $j$-th EDU in document $D_i$. This aggregation ensures that the document embedding $\mathbf{d}_i$ is primarily influenced by the most important EDUs, as determined by the attention weights ${a}_{(i,j,k)}$.


\subsection{Filtering and Ranking via Latent Query}
\label{sec:method_3}
As shown in Figure~\textcolor{blue}{\ref{fig:framework}} (b), after encoding EDUs and documents, the retrieval framework integrates filtering and ranking into a unified iterative process, represented by the filtering and ranking models, respectively. This process combines EDU-level salience estimation with document-level relevance scoring, and is performed within the EM algorithm, enabling the dynamic selection of latent queries and the updating of parameters in our retrieval framework.

\paragraph{\textbf{Filtering Model}}
The filtering model performs 
two tasks of selecting the most salient EDUs as latent queries and identifying the least salient EDUs for filtering.
Within this model, EDU embeddings are refined through cross-attention~\cite{9711309} with document embeddings, and then the salience score is estimated based on the refined EDU embeddings.
The cross-attention updates the EDU embeddings by aligning them with the document embeddings $\mathbf{D}$: 
\begin{equation}
    \mathbf{E} = \text{CrossAtt}(\mathbf{E}, \mathbf{D}, \mathbf{D}),
    \label{eq:updated_edu_embedding}
\end{equation}
where $\mathbf{E}=\{\mathbf{e}_{i,j}|E_{i,j}\in \mathcal{E}\}$ are all EDU embeddings which serve as a \textbf{query}{\footnote{{Please do not confuse the "query" in cross-attention with latent queries. The former refers to the query vector used in attention computation, while the latter serves as the selected EDUs for ranking.}}} 
in cross-attention, and $\mathbf{D}=\{\mathbf{d}_1,...,\mathbf{d}_n\}$ are all document embeddings which serve as the \textbf{key} and \textbf{value} in cross-attention. This cross-attention ensures that the updated embeddings $\mathbf{E}$ emphasize EDUs shared across multiple documents, facilitating a more focused evaluation of their salience.

Then, a MLP-based classifier is applied to calculate the salience score $s_{i,j}$ of the $j$-th EDU in document $D_i$ as:
\begin{equation}
s_{i,j} = \text{softmax}(\mathbf{W}_{\text{filter}} \cdot \mathbf{e}_{i,j} + \mathbf{b}_{\text{filter}}),
\label{eq:filtering}
\end{equation}
where $\mathbf{W}_{\text{filter}}$ and $\mathbf{b}_{\text{filter}} $ are trainable parameters. Based on these salience scores, the top-$k$ EDUs with the highest scores are selected as latent queries $\mathbf{Q}=\{\mathbf{q}_1, \mathbf{q}_2, \dots, \mathbf{q}_k\}$, which represent critical aspects of documents and serve as the guidance for document ranking.

\paragraph{\textbf{Ranking Model}}
The ranking model utilizes a multi-query ranking mechanism based on the latent queries selected by 
the EDU salience score.
Building on the document embeddings in Eq.~\textcolor{blue}{\ref{eq:document_embedding_pooling}}, the multi-query ranking evaluates the relevance score by considering the contribution of (all or selected) queries in $\mathbf{Q}$. For a given query $\mathbf{q} \in \mathbf{Q}$, the relevance score $r_{i,q}$ of document $D_{i}$ is calculated as:
\begin{equation}
r_{i,q} = \frac{\exp(\text{sim}(\mathbf{d}_i, \mathbf{q}))}{\sum_{j=1}^n \exp(\text{sim}(\mathbf{d}_j, \mathbf{q}))},
\end{equation}
where $\text{sim}(\cdot, \cdot)$ is a similarity function, {dot product in our setting,} that evaluates the alignment between document embedding $\mathbf{d}_i$ and query embedding $\mathbf{q}$. 
The query embedding is selected based on the EDU salience score and corresponds to the EDU embedding in Eq.~\textcolor{blue}{\ref{eq:updated_edu_embedding}}. This process is repeated for all queries
and the relevance score across all queries are combined to form an overall relevance score for each document:
\begin{equation}
r_{i} = \frac{1}{k} \sum_{q \in \mathbf{Q}} r_{i,q}.
\end{equation}

This aggregated score provides a unified measure of document importance across all queries, ensuring that the ranking reflects diverse perspectives, since multiple queries may reflect multiple aspects of documents. 



\paragraph{\textbf{Expectation-Maximization Optimization}}
The EM algorithm is an iterative method used for parameter optimization, particularly when the model depends on unobserved (latent) variables or when dealing with incomplete data~\cite{mclachlan2008algorithm}. {In our work, it is particularly well-suited as the latent queries correspond to the latent variables in the process.}
The EM algorithm alternates between the E-step and the M-step. 
In the E-step, we select the most salient EDUs as latent queries, according to the EDUs' salience score in the filtering model. 
In the M-step, we update the parameters of both filtering and ranking models, according to the filtering and ranking losses.

The filtering loss is served for query selection and EDU filtering, and is implemented as an EDU ranking problem based on the salience score in Eq.~\textcolor{blue}{\ref{eq:filtering}}. This loss ranks the most salient EDUs at the top and the least ones at the bottom. The top-ranked EDUs are selected as ``\textit{query}'' EDUs, while the bottom-ranked EDUs are filtered out during the retrieval process. 
From this, our approach focuses only on accurately ranking the top and bottom EDUs, without considering the order of the other EDUs. Therefore, the filtering loss is optimized by the Bayesian Personalized Ranking (BPR) loss~\cite{10.5555/1795114.1795167}:
\begin{equation}
    \fontsize{8}{14}{\mathcal{L}_{\text{filter}} = -\sum_{i \sim \mathcal{P}_q} \sum_{j \in \overline{\mathcal{P}}_q} \log \sigma(s_{i} - s_{j}) -\sum_{i \sim \overline{\mathcal{P}}_f} \sum_{j \in {\mathcal{P}}_f} \log \sigma(s_{i} - s_{j}),}
\end{equation}
where $\sigma$ is the sigmoid function, $s_{i}$ is the salience score of the $i$-th EDU from Eq.~\textcolor{blue}{\ref{eq:filtering}}, and $\mathcal{P}_q$ and $\mathcal{P}_f$ represent the top EDUs (query EDUs) and the bottom EDUs (filtered EDUs) in the ground-truth EDU ranking, respectively. The $\overline{\mathcal{P}_q}$ and $\overline{\mathcal{P}_f}$ indicate the complement set of query EDUs and filtered EDUs, respectively.

Similarly, the ranking loss assigns higher ranks to documents based on their alignment with all latent queries. 
We utilize the BPR loss for document ranking as well:
\begin{align}
    \mathcal{L}_{\text{rank}} = -\sum_{(i,j)\sim \mathcal{P}_g} \log\sigma\left(r_{i} - r_{j}\right),
\end{align}
where $(i,j)\sim \mathcal{P}_{g}$ denotes that document $i$ is ranked higher than document $j$ in the ground-truth ranking, 
and $r_{i}$ indicates the relevance score between document $i$ and all queries in $\mathbf{Q}$. 
The overall objective function integrates both ranking and filtering losses: 
\begin{equation}
\mathcal{L}_{\text{total}} = \mathcal{L}_{\text{rank}} + \lambda \mathcal{L}_{\text{filter}},
\label{eq:overall_objective}
\end{equation}
where $\lambda$ is the balance weight. 
During training, the EM algorithm iteratively refines both EDU representations $\mathbf{e}_{i,j}$ and document embeddings $\mathbf{d}_i$, ensuring mutual optimization of filtering and ranking.

At last, the E-step and the M-step are alternately repeated, with $\mathbf{e}_{i,j}$ and $\mathbf{d}_i$ being refined in each iteration. 
Since query embeddings share the same representation with EDU representation $\mathbf{e}_{i,j}$, the refined $\mathbf{e}_{i,j}$ guides the selection of salient EDUs, while $\mathbf{d}_i$ improves the ranking of documents based on their relevance to the queries. This alternative optimization ensures that both latent queries and document embeddings dynamically adapt to the content's salience, optimizing the final ranking.

\subsection{Creating Training Dataset for Retrieval}
\label{sec:method_1}
We detail the preparation of a training dataset for our retrieval framework in Figure~\textcolor{blue}{\ref{fig:framework}} (c). 
For both query selection and EDU filtering, we measure the salience score for each EDU by aligning the EDU with the reference summary. 
For document ranking, we create the rank for documents by the relevance between each document and the reference summary.
These provide the ground truth (oracle) for filtering and ranking, ensuring the dataset aligns with the requirement of our framework and supports effective supervision.

\paragraph{\textbf{EDU Scoring}}
EDU-level scores are determined by evaluating the relevance of each segmented EDU against the reference summary. Using the EDU spans from segmentation annotation, we embed both the EDUs and the reference summary into a shared semantic space using a pre-trained model (multi-qa-mpnet-base-cos-v1) designed for Semantic Search~\cite{reimers-2019-sentence-bert}.
The relevance of each EDU is calculated as the cosine similarity between its embedding and the embedding of the reference summary. These scores reflect the salience of the EDUs within the overall context and are leveraged to supervise the EDU ranking process, including both latent queries selection and EDU filtering.

\paragraph{\textbf{Document Ranking}}
To establish document ranking, we calculate relevance scores between each document and the corresponding reference summary using the same pre-trained model used for EDU scoring. Each document is represented as a collection of EDU embeddings, and the overall document relevance score is computed by aggregating the cosine similarity scores of its EDUs with the gold summary. Documents are ranked based on these aggregated scores, with higher-ranked documents prioritized for inputs. 

\begin{table}[ht]
\centering
\caption{The statistics of the datasets. 
{$*$ denotes that we utilized the dataset provided by PRIMERA~\cite{DBLP:conf/acl/XiaoBCC22}}.}
\vspace{-5pt}
\setlength{\tabcolsep}{0.8mm}
\resizebox{0.47 \textwidth}{!}
{
\begin{tabular}{lcccc}
\hline
\textbf{Dataset} & \textbf{Train/Dev./Test} & \textbf{Avg. Docs} & \textbf{Len$_\text{src}$} & \textbf{Len$_\text{tgt}$} \\
\hline
Multi-News \cite{fabbri-etal-2019-multi} & 44,972/5,622/5,622 & 2.8 & 1793 & 217 \\
Multi-Xscience \cite{lu-etal-2020-multi-xscience} & 30,369/5,066/5,093 & 4.4 & 700 & 105 \\
Wikisum* \cite{cohen-etal-2021-wikisum} & 38,219/38,144/3,200 & 40 & 2238 & 113 \\
WCEP-10 \cite{gholipour-ghalandari-etal-2020-large} & 8,158/1,020/1,020 & 9.1 & 3866 & 28 \\
\hline
\end{tabular}}
\label{tab:dataset}
\vspace{-8pt}
\end{table}

\subsection{Generating Summaries in MDS}
\label{sec:method_4}
As shown in Figure~\textcolor{blue}{\ref{fig:framework}} (d), after obtaining the well-trained retrieval framework, we infer the EDU filtering scores and document ranking scores. 
We then apply filtering and ranking to generate summaries. 
Specifically, the retrieval model takes documents as input and generates relevance scores for documents and salience scores for their EDUs. Using these inferred relevance scores, we rank the documents, ensuring that the least important documents are truncated. 
Instead of simply dropping the last tokens during truncation, we filter out the lowest-ranked EDUs by the salience scores to fit the context length of downstream summarizer.

\begin{table*}[ht]
\centering
\caption{Comparison in different supervised 
settings. The number in parentheses is
the token length limit for the model. $-$ denotes that the improvement is not significant (p>=0.05) compared
with the original score, using the paired bootstrap
resampling method \cite{koehn-2004-statistical}.
}
\vspace{-5pt}
\setlength{\tabcolsep}{1.5mm}
{
{ 
\begin{tabular}{l@{\hspace{2mm}}:c|lll|lll|lll|lll}
\toprule
\multirow{2}{*}{\textbf{Model}} & \multirow{2}{*}{\textbf{Settings}} & \multicolumn{3}{c|}{\textbf{Multi-News}} & \multicolumn{3}{c|}{\textbf{Multi-XScience}} & \multicolumn{3}{c|}{\textbf{Wikisum}} &  \multicolumn{3}{c}{\textbf{WCEP-10}}\\
 & & \multicolumn{1}{c}{R-1} & \multicolumn{1}{c}{R-2} & \multicolumn{1}{c|}{R-L}  & \multicolumn{1}{c}{R-1} & \multicolumn{1}{c}{R-2} & \multicolumn{1}{c|}{R-L} & \multicolumn{1}{c}{R-1} & \multicolumn{1}{c}{R-2} & \multicolumn{1}{c|}{R-L} & \multicolumn{1}{c}{R-1} & \multicolumn{1}{c}{R-2} & \multicolumn{1}{c}{R-L} \\
\midrule
\midrule
\multirow{2}{*}{BartGraphSum(4050)}
 & original & 46.49 & 18.70 &  23.71 & 29.13 & 6.49 & 16.98 & 39.55 & 23.56 & 32.11 & 44.89 & 23.56 & 36.37 \\
 & retrieval & \textbf{47.73} & \textbf{19.38} & \textbf{24.66} &  {29.67}$^{-}$ & {6.78}$^{-}$ & {17.06}$^{-}$ &  \textbf{41.33} & \textbf{25.50} & \textbf{34.01} & \textbf{46.29} & \textbf{24.64} & \textbf{37.70}
 \\
\midrule
\multirow{2}{*}{BART(1024)}
 & original & 44.87 & 16.93 & 22.21 & 22.21 & 4.91 & 13.70 & 39.66 & 22.54 & 32.26 & 43.24 & 21.96 & 34.66 \\
 & retrieval & \textbf{46.46} & \textbf{18.52} & \textbf{23.83} &  \textbf{24.75} & \textbf{5.69} & \textbf{14.79} &  \textbf{42.22} & \textbf{24.98} & \textbf{33.95} & \textbf{46.25} & \textbf{23.92} & \textbf{37.14}
 \\
\midrule
\multirow{2}{*}{PEGASUS(512)}
 & original & 47.70 & 18.36 & 23.62 & 28.73 & 5.21 & 15.99 & 31.21 & 16.46 & 25.68 & 42.43 & 17.33 & 32.35 \\
 & retrieval & \textbf{48.97} & \textbf{19.21} & \textbf{24.67} &  \textbf{30.24} & \textbf{5.78} & \textbf{16.93} &  \textbf{35.85} & \textbf{18.80} & \textbf{27.48} & \textbf{45.42} & \textbf{23.66} & \textbf{36.34}
 \\
\midrule
\multirow{2}{*}{PRIMERA(4096)}
 & original & 49.84 & 19.97 & 24.88 & 32.67 & 7.12 & 18.01 & 42.42 & 25.32 & 35.01 & 46.08 & 24.15 & 36.71 \\
 & retrieval & \textbf{50.27} & \textbf{20.38} & \textbf{25.33} &  \textbf{34.73} & \textbf{7.44} & \textbf{18.29} &  \textbf{44.53} & \textbf{27.10} & \textbf{36.72} & \textbf{47.33} & \textbf{25.08} & \textbf{37.87}
 \\
\midrule
\multirow{2}{*}{LLaMA 3.1-8B(4096)}
 & original & 43.25 & 14.24 & 20.74 & 31.78 & 6.16 & 17.46 & 41.64 & 24.65 & 34.08 & 43.50 & 20.77 & 34.01 \\
 & retrieval & \textbf{47.88} & \textbf{17.82} & \textbf{23.58} &  {32.49}$^-$ & \textbf{6.83} & \textbf{17.95} &  \textbf{43.51} & \textbf{26.52} & \textbf{36.23} & \textbf{45.09} & \textbf{22.40} & \textbf{35.46}
 \\
\midrule
\multirow{2}{*}{LLaMA 3.2-1B(4096)}
 & original & 42.31 & 12.94 & 20.26 & 29.67 & 5.33 & 16.56 & 31.10 & 15.45 & 23.84 & 40.33 & 17.96 & 31.20 \\
 & retrieval & \textbf{46.37} & \textbf{16.46} & \textbf{22.88} &  \textbf{30.73} & \textbf{5.76} & \textbf{17.13} &  \textbf{36.81} & \textbf{19.35} & \textbf{28.90} & \textbf{41.44} & \textbf{19.46} & \textbf{32.83}
 \\
 \midrule
\multirow{2}{*}{{\shortstack{StableLm-Zephyr-3B  \\ (2048)}}}
 & original & 43.36 & 13.10 & 20.35 & 32.45 & 6.25 & 17.78 & 39.57 & 20.90 & 31.18 & 42.56 & 19.62 & 33.03 \\
 & retrieval & \textbf{46.84} & \textbf{16.31} & \textbf{22.64} & 32.63$^-$ & 6.34$^-$ & 17.84$^-$ &  \textbf{40.56} & \textbf{21.93} & \textbf{31.98} & 43.16$^-$ & 19.98$^-$ & 33.83$^-$
 \\
\bottomrule
\end{tabular}}}
\label{tab:sup_rouge}
\end{table*}

\section{Experiments}

\subsection{Settings}\label{settings}

\hspace{1.2em}\textit{\textbf{Evaluation Datasets.}} 
We evaluated our approach on four multi-document summarization datasets from various domains (News, Scientific literature, and Wikipedia) and adopted the provided data split from the datasets. See Table~\textcolor{blue}{\ref{tab:dataset}} for detailed dataset statistics. 

\vspace{5pt}
\textit{\textbf{Base Summarization Models.}} 
We evaluated our retrieval framework to understand its effectiveness using various models
and configurations. The models include BartGraphSum \cite{DBLP:conf/naacl/PasunuruLBRD21}, BART \cite{lewis-etal-2020-bart}, PEGASUS \cite{pmlr-v119-zhang20ae}, PRIMERA \cite{DBLP:conf/acl/XiaoBCC22}, LLaMA variants (3.1-8B\footnote{\textcolor{blue}{\url{https://huggingface.co/meta-llama/Llama-3.1-8B-Instruct}}} and 3.2-1B\footnote{\textcolor{blue}{\url{https://huggingface.co/meta-llama/Llama-3.2-1B-Instruct}}}), and StableLM-Zephyr-3B\footnote{\textcolor{blue}{\url{https://huggingface.co/stabilityai/stablelm-zephyr-3b}}}. BartGraphSum is a graph-enhanced model that incorporates an IE graph, constructed from Open Information Extraction (OIE) triplets, to improve summarization quality. BART is a transformer-based denoising autoencoder designed for sequence-to-sequence tasks. PEGASUS employs the Gap Sentence Generation (GSG) pre-training objective, where key sentences are masked and reconstructed, making it well-suited for summarization. PRIMERA is built on the LED \cite{beltagy2020longformer} architecture and uses pyramid-based masked sentence pre-training to effectively aggregate information across multiple documents. The LLaMA models include the resource-intensive 3.1-8B variant and the efficient 3.2-1B model. StableLM-Zephyr-3B offers a lightweight LLM. For fair comparison, we used the same input context as PRIMERA when evaluating LLaMA variants, as LLaMA-3.1 supports a context length of up to 128k tokens. 
The LLaMA and StableLM are fine-tuned with Low-Rank Adapters (LoRA) \cite{DBLP:journals/corr/abs-2106-09685}.

\vspace{5pt}
\textit{\textbf{Evaluation Metrics.}} 
{For summarization evaluation,} we followed previous work~\cite{DBLP:conf/acl/XiaoBCC22} and used ROUGE \cite{lin-2004-rouge} scores {(R-1, R-2, and R-L)}, which are the standard evaluation metrics.
\footnote{We used \textcolor{blue}{\url{https://github.com/google-research/google-research/tree/master/rouge}} with default stemmer settings.}
{For both query selection and filtering evaluations}, we used Precision@K (Pre@K), which measures the proportion of salient EDUs among the top-K ranked EDUs for query selection, and the proportion of irrelevant EDUs among the bottom-K ranked EDUs for filtering.
For document ranking evaluation, {we used NDCG@K~\cite{DBLP:journals/tois/JarvelinK02}, MRR\_1st~\cite{DBLP:conf/lrec/RadevQWF02}, and MRR\_2nd, where NDCG@K assesses the quality of ranking by considering the positions of documents in the top-K list of the results, and MRR\_1st/MRR\_2nd focuses on the positions of the top-1/top-2 ground-truth documents in the list of the results.
}

\vspace{5pt}
\textit{{\textbf{Original Truncation.}}}
{For the original baseline models, we used their default truncation strategy, which truncates each input document evenly to fit the context length, as in \cite{fabbri-etal-2019-multi,DBLP:conf/acl/XiaoBCC22}.  
Specifically, if there are $n$ documents with a total length of $l$ tokens, the baseline models select $l/n$ tokens from each document and concatenate the truncated documents as an input. 
Since LLaMA and StableLM are not specifically designed for summarization tasks, we apply the same truncation strategy as other baseline models for consistency}.

\vspace{5pt}
\textit{{\textbf{Hyperparameter Settings.}}} 
{We performed a grid search to find the best hyperparameters on the validation dataset. 
Unless otherwise specified, we set the chunk size $c$ to 1024 tokens, the filtering loss weight $\lambda$ to $1.0$, and the query number $k$ to $10$, ensuring consistent comparison across models and configurations. 
Additionally, we used the Adam optimizer with a learning rate of $3e-5$ and batch size of $16$ for BartGraphSum, BART, PEGASUS, and PRIMERA, while a learning rate of $2e-4$ and batch size of $8$ for LLaMA and StableLM}.

\subsection{Evaluation in Fully Supervised Settings}
Table \textcolor{blue}{\ref{tab:sup_rouge}} shows consistent improvements achieved by our framework in the ROUGE metrics across various datasets 
and model configurations. Our retrieval framework achieves a significant performance gain in all datasets compared to the original base models. 
Specifically, 
our framework achieves an average gain of $+0.43$ on Multi-News, $+0.95$ on Multi-XScience, $+1.85$ on Wikisum, and $+1.11$ on WCEP-10 {in PRIMERA}. This demonstrates that our retrieval effectively filters out irrelevant information and prioritizes the salient content, enhancing overall summarization quality. 
Furthermore, it excels in settings with shorter context lengths (512 tokens), as seen with PEGASUS. The results for {BART} and PEGASUS across all MDS datasets highlight how our retrieval framework
efficiently complements
its pre-training strategy. Even LLMs such as the LLaMA family and StableLM show notable gains, emphasizing the universal importance of prioritizing salient information and minimizing irrelevant content, regardless of scale. These findings underscore the robustness of our retrieval {framework} in refining input quality and its versatility in enhancing diverse architectures.

\begin{table*}[h]
\centering
\caption{Comparison in different few-shot evaluation settings. The notations are the same as those in Table~\textcolor{blue}{\ref{tab:sup_rouge}}.
}
\vspace{-5pt}
\setlength{\tabcolsep}{1.5mm}
{
{ 
\begin{tabular}{l@{\hspace{2mm}}:c|lll|lll|lll|lll}
\toprule
\multirow{2}{*}{\textbf{Model}} & \multirow{2}{*}{\textbf{Settings}} & \multicolumn{3}{c|}{\textbf{Multi-News}} & \multicolumn{3}{c|}{\textbf{Multi-XScience}} & \multicolumn{3}{c|}{\textbf{Wikisum}} &  \multicolumn{3}{c}{\textbf{WCEP-10}}\\
 & & \multicolumn{1}{c}{R-1} & \multicolumn{1}{c}{R-2} & \multicolumn{1}{c|}{R-L}  & \multicolumn{1}{c}{R-1} & \multicolumn{1}{c}{R-2} & \multicolumn{1}{c|}{R-L} & \multicolumn{1}{c}{R-1} & \multicolumn{1}{c}{R-2} & \multicolumn{1}{c|}{R-L} & \multicolumn{1}{c}{R-1} & \multicolumn{1}{c}{R-2} & \multicolumn{1}{c}{R-L} \\
\midrule
\midrule
\multirow{2}{*}{BartGraphSum(4050)}
 & original & 41.26 & 14.99 & 20.12 & 25.76 & 4.89 & 14.65 & 31.80 & 14.29 & 24.63 & 37.87 & 17.00 & 29.09 \\
 & retrieval & \textbf{42.71} & \textbf{15.56} & \textbf{21.49} &  {26.48}$^-$ & {4.94}$^-$ & \textbf{15.77} &  \textbf{35.36} & \textbf{18.18} & \textbf{27.84} & {38.03}$^-$ & {16.97}$^-$ & {29.18}$^-$
 \\
\midrule
\multirow{2}{*}{BART(1024)}
 & original & 41.23 & 15.81 & 20.94 & 21.56 & {4.98} & 11.34 & 28.20 & 10.99 & 20.76 & 35.66 & 15.21 & 26.47 \\
 & retrieval & \textbf{42.68} & {16.18}$^-$ & {21.47}$^-$ &  \textbf{23.29} & 4.93$^-$ & \textbf{12.40} &  \textbf{31.21} & \textbf{13.07} & \textbf{23.46} & \textbf{38.79} & \textbf{17.72} & \textbf{29.56}
 \\
\midrule
\multirow{2}{*}{PEGASUS(512)}
 & original & 42.39 & 13.20 & 20.70 & 27.40 & 4.76 & 15.05 & 23.89 & 5.51 & 13.90 & 38.18 & 18.94 & 29.53 \\
 & retrieval & \textbf{43.78} & \textbf{14.13} & \textbf{22.51} &  \textbf{28.69} & \textbf{5.01} & {15.27}$^-$ &  \textbf{31.36} & \textbf{11.76} & \textbf{19.67} & \textbf{39.54} & \textbf{19.88} & \textbf{30.49}
 \\
\midrule
\multirow{2}{*}{PRIMERA(4096)}
 & original & 46.10 & 16.80 & 22.63 & 29.53 & 5.31 & 15.27 & 36.48 & 17.85 & 28.43 & 40.66 & 18.64 & 31.34 \\
 & retrieval & \textbf{47.38} & \textbf{17.34} & \textbf{23.09} &  \textbf{31.21} & \textbf{5.62} & \textbf{16.20} &  \textbf{39.29} & \textbf{20.17} & \textbf{30.60} & \textbf{42.46} & \textbf{19.72} & \textbf{32.36}
 \\
\midrule
\multirow{2}{*}{LLaMA 3.1-8B(4096)}
 & original & 42.16 & 13.13 & 20.09 & 30.44 & 5.58 & 16.85 & 40.72 & 22.66 & 32.60 & 40.96 & 18.62 & 31.44 \\
 & retrieval & \textbf{45.56} & \textbf{16.01} & \textbf{21.97} &  \textbf{31.01} & \textbf{5.83} & \textbf{17.09} &  \textbf{41.85} & \textbf{23.93} & \textbf{33.92} & \textbf{43.36} & \textbf{19.81} & \textbf{33.35}
 \\
\midrule
\multirow{2}{*}{LLaMA 3.2-1B(4096)}
 & original & 38.16 & 12.02 & 18.57 & 27.64 & 4.15 & 14.48 & 29.79 & 15.64 & 25.33 & 36.62 & 16.68 & 29.41 \\
 & retrieval & \textbf{39.23} & \textbf{13.01} & \textbf{19.77} &  \textbf{28.53} & \textbf{4.57} & \textbf{15.04} &  \textbf{32.58} & \textbf{16.52} & \textbf{26.81} & \textbf{38.90} & \textbf{17.85} & \textbf{30.91}
 \\
 \midrule
\multirow{2}{*}{\shortstack{StableLm-Zephyr-3B  \\ (2048)}}
 & original & 41.58 & 12.45 & 19.71 & 30.20 & 5.28 & 16.55 & 36.26 & 17.10 & 27.41 & 40.58 & 17.92 & 31.33 \\
 & retrieval & \textbf{45.09} & \textbf{14.54} & \textbf{21.31} &  30.36$^-$ & 5.33$^-$ & 16.67$^-$ &  \textbf{37.26} & \textbf{18.45} & \textbf{28.28} & 41.28$^-$ & 18.28$^-$ & 32.23$^-$
 \\

\bottomrule
\end{tabular}}}
\label{tab:few_shot_rouge}
\end{table*}

\subsection{Evaluation in Few-shot Settings}
We further evaluated the retrieval framework under few-shot settings to understand its adaptability with limited supervision. We used only 1\% of the available training data for few-shot evaluation. Table~\textcolor{blue}{\ref{tab:few_shot_rouge}} presents the results. 
Our retrieval framework demonstrates consistent performance improvements across most datasets and models.
Our retrieval {framework} shows impressive results on the Wikisum dataset, which is characterized by its highly fragmented input structure, with an average of 40 source documents per summary. Notably, PEGASUS achieves a significant $+7.47$ point improvement in ROUGE-2, demonstrating the framework’s ability to filter irrelevant content and prioritize salient information effectively. This capability allows models to generate informative and concise summaries, even in the challenging few-shot settings, where limited supervision further complicates content selection. These results highlight that our retrieval {framework} enhances performance not only in full-supervision 
scenarios but also in limited supervision ones. 
Its ability to handle dispersed inputs underscores its versatility and practical value in real-world summarization tasks.

\begin{figure}[h]
	\centering
        \includegraphics[scale=0.43]{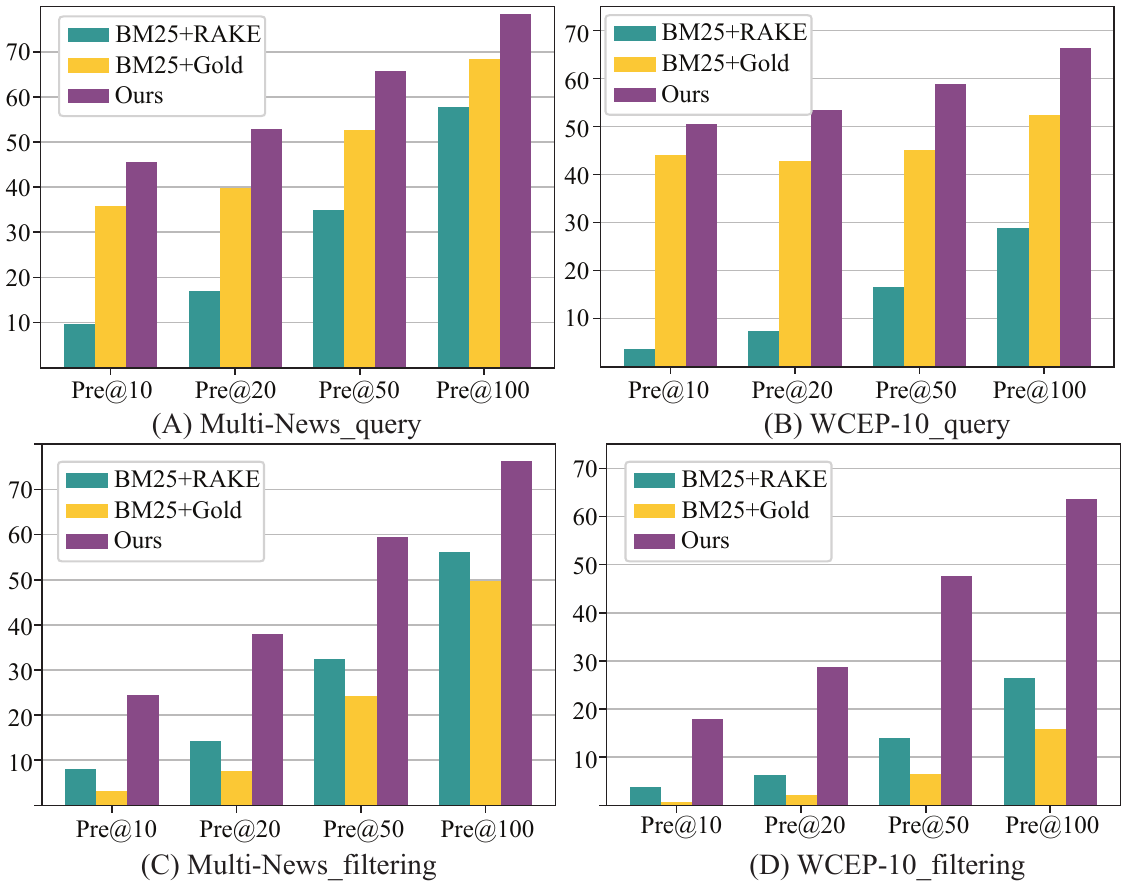}
    \vspace{-5pt}
	\caption{Precision@K (K = 10, 20, 50, 100) for query selection and filtering. (A) (B): the proportion of the top-K selected EDUs for query selection that align with the most salient EDUs. (C) (D): the proportion of the bottom-K selected EDUs for filtering that align with the least salient EDUs. }
    \label{fig:query_acc}
    \vspace{-5pt}
\end{figure}

\subsection{In-depth Analysis}
To evaluate the accuracy of our retrieval framework, we analyzed its performance on two key tasks: {EDU selection} and document ranking. For {EDU selection, which includes both query selection and EDU filtering,} we compared it against BM25 \cite{10.1561/1500000019}, a bag-of-words retrieval function that ranks content based on the occurrence of query terms. For document ranking, we used BM25 \cite{10.1561/1500000019} and DYLE \cite{DBLP:conf/acl/MaoWN0Z0DZAR22} as baselines. DYLE\footnote{We used the arXiv retriever checkpoint provided by the authors.} jointly trains a retriever and a generator, treating retrieved sentences as latent variables to guide dynamic attention and highlight salient information for long-document summarization. We focused on the retriever component of DYLE to compare it with our module in the effectiveness in document ranking. Since both BM25 and DYLE require a query as an input, we used RAKE (Rapid Automatic Keyword Extraction) \cite{rose2010automatic}, an unsupervised keyword extraction algorithm, to simulate queries.
{For the EDU selection task, we used the ranked EDUs based on the EDU scoring from Section \textcolor{blue}{\ref{sec:method_1}} as the ground truth. Specifically, the top-k EDUs were used as the ground truth for Precision@K in query selection and the bottom-K EDUs were used in EDU filtering. For the document ranking task, we used the document ranking labels from Section \textcolor{blue}{\ref{sec:method_1}} as the ground truth.}


\begin{figure}[h]
	\centering
        \includegraphics[scale=0.43]{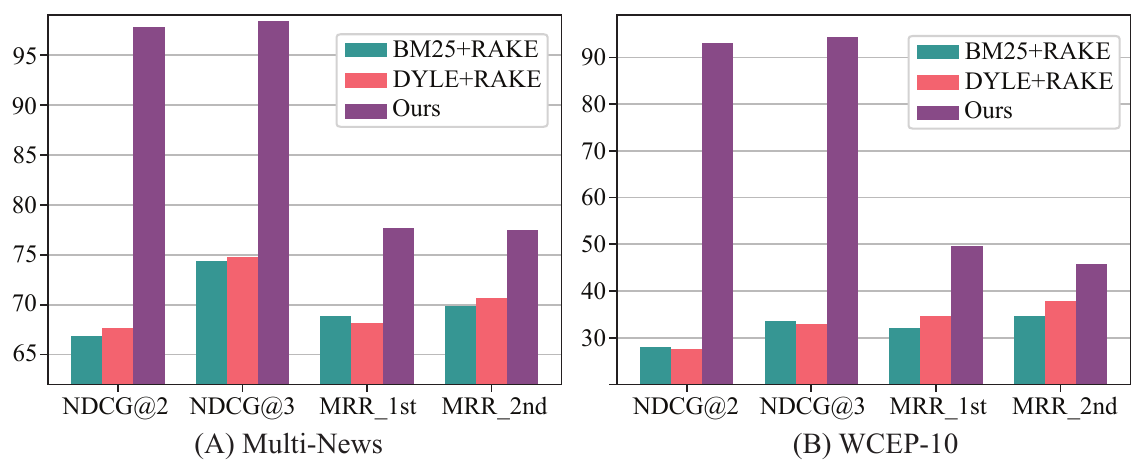}
        \vspace{-5pt}
	\caption{Document ranking accuracy. NDCG@K evaluates how effectively the most {relevant} documents are ranked within the top K positions. MRR\_1st and MRR\_2rd measure the average rank accuracy of the most and second most {relevant} documents, respectively. 
    (A): Multi-News. (B): WCEP-10.
    }
    \vspace{-10pt}
    \label{fig:ranking_acc}
\end{figure}

For query selection, we evaluated how effectively the retrieval {framework} identifies the most {salient} EDUs.
Unlike BM25, which requires an explicit query to retrieve content, our retrieval {framework} dynamically selects EDUs directly from input documents. To ensure fair comparison, we constructed two variants of BM25: (1) BM25+RAKE, where the query is derived using RAKE to generate realistic and robust baselines; and (2) BM25+Gold, where the query is the EDU with the highest relevance score to the reference summary, as defined in Section \textcolor{blue}{\ref{sec:method_1}}. 
As shown in Figures~\textcolor{blue}{\ref{fig:query_acc}} (A) and (B), our framework outperforms BM25+RAKE and BM25+Gold across all thresholds, demonstrating its ability to better align with salient EDUs.
Remarkably, even when BM25+Gold utilizes {the most relevant} 
EDUs to the reference summary as a query, our retrieval {framework} significantly outperforms it across all thresholds. 
Similarly, for {EDU filtering, we evaluated how effectively the retrieval {framework} identifies the least salient EDUs}
(Figures~\textcolor{blue}{\ref{fig:query_acc}} (C) and (D)). Our module exhibits superior performance, effectively identifying least salient content. We attribute the lower performance of BM25+Gold compared to BM25+RAKE to differences in their query formulation: Gold ensures high specificity by aligning with the reference summary but lacks contextual breadth, whereas RAKE captures broader document context through co-occurrence patterns, albeit with less specificity.
These results indicate that our module is not only adept at selecting {salient} EDUs but also proficient in filtering extraneous information, ensuring 
{a context-aware retrieval that is independent of the human-written queries.}
Building on these results, we next evaluated the document ranking accuracy of our retrieval {framework}.
We compared our module against BM25+RAKE and DYLE+RAKE baselines. As shown in Figure~\textcolor{blue}{\ref{fig:ranking_acc}}, our framework consistently achieves higher NDCG@K and MRR scores across datasets. On Multi-News, for instance, it surpasses the baselines in both metrics, ranking the most salient documents at the top. Furthermore, on more challenging datasets like WCEP-10, our retrieval {framework} demonstrates robust performance with significant improvements over the baselines. These results emphasize the module's capability to align with reference content while reducing dependence on pre-defined explicit queries.

\subsection{Human Evaluation}
To assess the quality of generated summaries, we conducted a human evaluation on a randomly sampled 100 data points from 
the Multi-News test dataset. The evaluation was conducted using Amazon Mechanical Turk,\footnote{\textcolor{blue}{\url{https://www.mturk.com/}}} with 10 evaluators holding at least a US high school diploma and a bachelor's degree. 
Evaluators assess the overall quality of summaries on a Likert scale from 1 to 5 (5 being the highest) across three criteria, {following the evaluation protocol of GraphSum~\cite{li-etal-2020-leveraging-graph}}: (1) \textit{Informativeness}: does the summary convey important facts of the input? (2) \textit{Fluency}: is the summary fluent and grammatical? (3) \textit{Succinctness}: does the summary avoid repeating information?
The results of the evaluation are presented in Table~\textcolor{blue}{\ref{tab:human_eval}}. 
Our retrieval {framework} shows its effectiveness in improving the quality of generated summaries. Specifically, it outperforms the original {truncation} of the base model {PRIMERA} 
in informativeness and fluency, while maintaining comparable performance in succinctness. These results highlight the capability of our retrieval {framework} to generate summaries that are not only more informative {and fluent but also concise.}

\begin{table}[h]
\centering
\caption{Human Evaluation of our retrieval {framework}. The notations are the same as those in Table~\textcolor{blue}{\ref{tab:sup_rouge}}.
}
\vspace{-5pt}
\setlength{\tabcolsep}{2.2mm}
{
\begin{tabular}{lccc}
\toprule
\textbf{Settings} & \textbf{Informativeness} & \textbf{Fluency} & \textbf{Succinctness}  \\
\midrule
\textbf{original} & 4.03 & 4.20 & 4.30 \\
\textbf{retrieval}& \textbf{4.16} & \textbf{4.38} & \,\,\,\,4.31$^{-}$  \\
\bottomrule
\end{tabular}
\label{tab:human_eval}}
\vspace{-5pt}
\end{table}

\subsection{{Ablation Study} and Parameter Analysis}

To further evaluate the effectiveness of our retrieval {framework}, we conducted an ablation study and parameter analysis.
These experiments were conducted using PRIMERA.\footnote{{While we show the parameter analysis only on the Multi-News dataset, other datasets also exhibit the same parameter trends.}} 

\paragraph{\textbf{Ablation Study}}
To investigate the effectiveness of ranking and filtering in our retrieval {framework}, 
we applied the following variants: 
(1) ``w/o rank'': we used only the EDU filtering in our truncation strategy, with using random permutation for the rank of documents.
(2) ``w/o filter'': we used only the document ranking, 
where documents were ranked, but the last tokens were dropped in stead of filtering irrelevant EDUs.
(3) ``w/o both'': both document ranking and filtering were removed, where randomly permuted documents were concatenated and the last tokens were dropped to fit the length limitation.
Experimental results in Table~\textcolor{blue}{\ref{tab:ablation}} show that ``w/o rank'' and ``w/o filter'' result in a measurable decrease across metrics, revealing that both ranking and filtering benefit our retrieval {framework} in summarization. Furthermore, ``w/o both'' has the lowest performance, demonstrating the complementary nature of ranking and filtering in enhancing the retrieval {framework}.\footnote{{Please 
note that our truncation strategy variant, {``w/o both''}, is different from the original truncation in PRIMERA, and the results of {``w/o both''} in Table \textcolor{blue}{\ref{tab:ablation}} do not align with {``original''} in Table \textcolor{blue}{\ref{tab:sup_rouge}}. Specifically, the original truncation evenly truncates each document. In contrast, the {``w/o both''} variant concatenates all randomly permuted documents and drops the last tokens, which truncates only the last ranked documents.}}

\begin{table}[h]
\centering
\caption{Ablation study of our retrieval {framework}.}
\vspace{-5pt}
{
\resizebox{0.46 \textwidth}{!}
{ 
\begin{tabular}{l@{\hspace{2mm}}|ccl|ccc}
\toprule
\multirow{2}{*}{\textbf{Settings}} & \multicolumn{3}{c|}{\textbf{Multi-News}} & \multicolumn{3}{c}{\textbf{WCEP-10}} \\
 & R-1 & R-2 & R-L & R-1 & R-2 & R-L \\
\midrule
\midrule
Ours & 50.27 & 20.38 & 25.33 & 47.33 & 25.08 & 37.87 \\
w/o rank & 49.75 & 20.25 & 25.15 & 46.08 & 24.26 & 36.74 \\
w/o filter & 49.95 & 20.26 & 25.22$^-$ & 46.75 & 24.66 & 37.23 \\
w/o both & 49.55 & 20.14 & 25.07 & 44.92 & 22.59 & 35.01 \\
\bottomrule
\end{tabular}}}
\label{tab:ablation}
\vspace{-10pt}
\end{table}

\begin{figure}[h]
	\centering
        \includegraphics[scale=0.365]{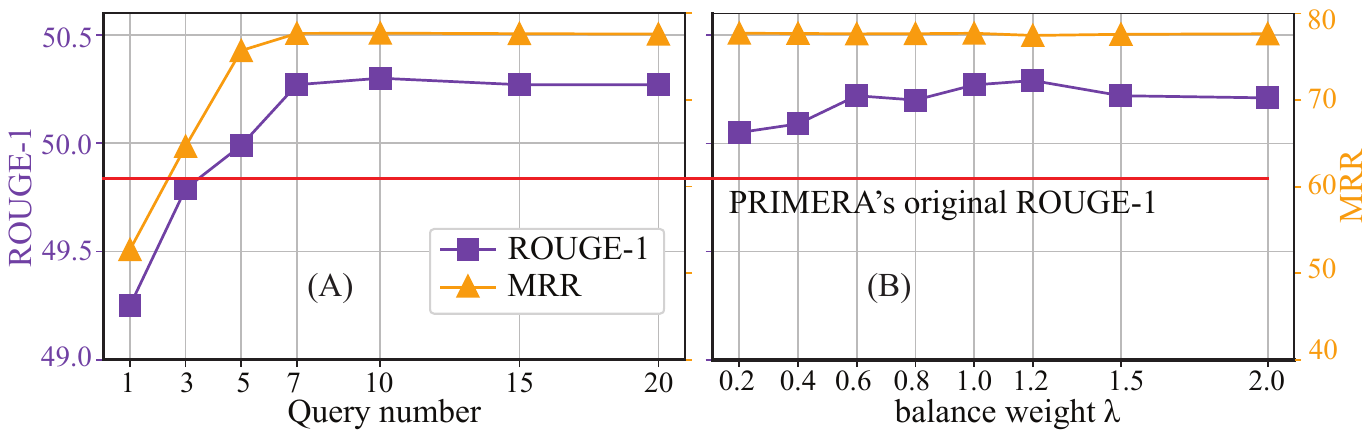}
	\vspace{-15pt}
    \caption{ROUGE score and document accuracy with different parameter settings. (A): varying the query number. (B): varying the loss balance weight $\lambda$. }
    \vspace{-10pt}
    \label{fig:debiasing_effect}
\end{figure}

\paragraph{\textbf{Parameter Analysis}}
We evaluated the parameter sensitivity by varying the target hyperparameter on the validation dataset while keeping other hyperparameters fixed. 
Specifically, we investigated the influence of query numbers and loss balance weight $\lambda$ in the ranking accuracy and summarization performance. 
We found that the number of latent queries significantly impacts the ranking accuracy of documents. As shown in Figure~\textcolor{blue}{\ref{fig:debiasing_effect}} (A), MRR rapidly increases as the query number grows from 1 to 7. This suggests that a small number of queries is insufficient to capture the diversity of salient information in multi-document summarization. The ROUGE-1 score also follows a similar trend, saturating 
at around 10 queries, which shows the highest performance.
We also identified that the balance weight $\lambda$ in the loss function has minimal influence on both ranking and summarization performance. Figure~\textcolor{blue}{\ref{fig:debiasing_effect}} (B) illustrates that varying $\lambda$ from 0.2 to 2.0 results in only marginal changes in both the ROUGE-1 score and MRR. This demonstrates that the framework is robust to changes in the balance weight $\lambda$, allowing for flexibility in tuning without significantly affecting performance. The consistent results across different $\lambda$ values highlight the stability of the proposed model.

\subsection{Case Study}
\begin{figure}[t]
	\centering
        \includegraphics[scale=0.57]{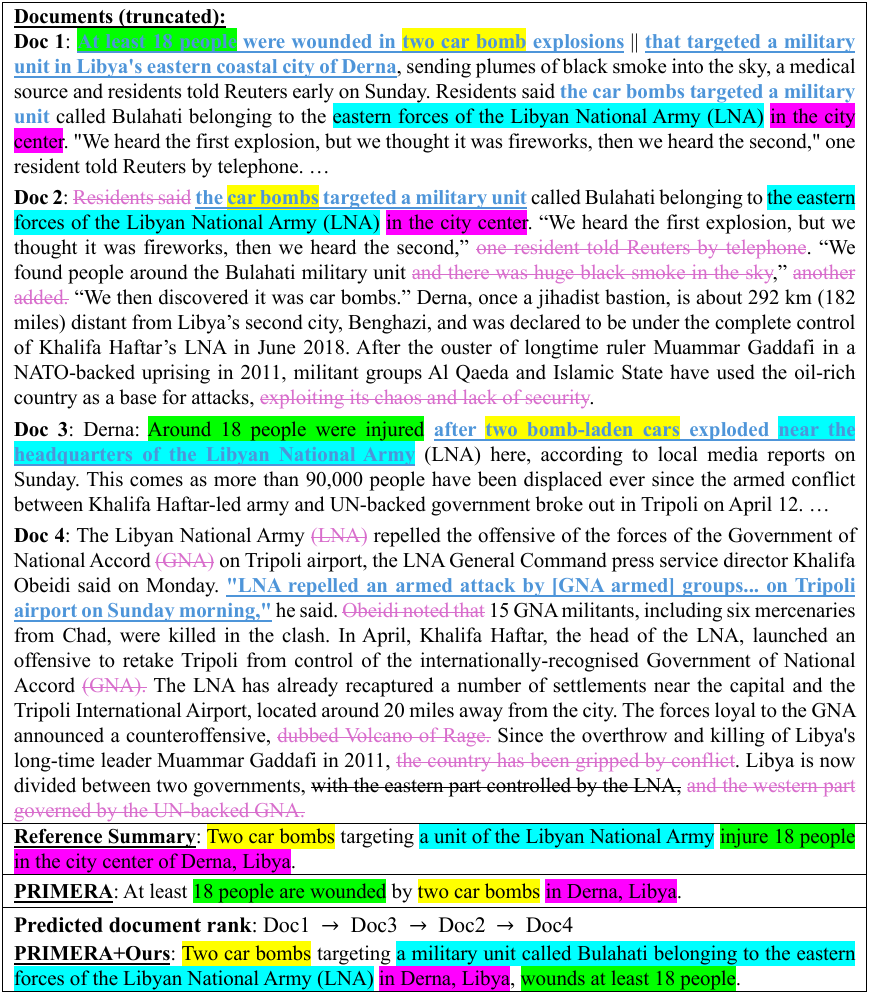}
        \vspace{-5pt}
	\caption{Example of generated summaries, latent query selection, ranking, and filtering. }
    \label{fig:ranking_filtering_case}
    \vspace{-8pt}
\end{figure}

Figure~\textcolor{blue}{\ref{fig:ranking_filtering_case}} illustrates an example of the generated summary using PRIMERA with and without our retrieval {framework}. In the figure, key information is highlighted in the same color, \textit{e.g.,}  \sethlcolor{yellow}\hl{Two car bombs}, selected queries are underlined in blue, \textit{e.g.,} \textcolor{lightblue}{\underline{At least 18...}}, and filtered EDUs are struck through in pink, \textit{e.g.,} {\textcolor{palepink}{\sout{(GNA)}}}. Our retrieval {framework} ranks Document 1 as the highest for its rich key information and Document 4 as the lowest for its irrelevance. 
By incorporating our retrieval {framework} into the truncation strategy, we filter out EDUs from the lower-ranked documents (Documents 2 and 4), effectively removing repetitive and irrelevant content. 
Additionally, we observe that selected queries, identified by our retrieval {framework}, align closely with the reference summary, such as \textit{"two car bombs targeting a military unit in Derna."}


\section{Conclusion}
{In this work, we aimed to propose a retrieval model that tackles the context length limitation in MDS models. To achieve this goal, 
}
{we} proposed a novel retrieval framework for multi-document summarization that integrates both {query selection, document ranking and EDU filtering} into an iterative Expectation-Maximization process. In our evaluation across diverse datasets and base models, the framework consistently improved ROUGE and ranking-specific metrics such as precision and MRR. The analysis demonstrated its scalability, flexibility, and potential for integration into existing pipelines. {In future work, we aim to enhance our framework's efficiency when it is applied to large-scale, real-world multi-document summarization scenarios.}


\bibliographystyle{ACM-Reference-Format}
\balance
\bibliography{main}



\end{document}